# Automatic Detection of Online Jihadist Hate Speech

Tom De Smedt, Guy De Pauw and Pieter Van Ostaeyen

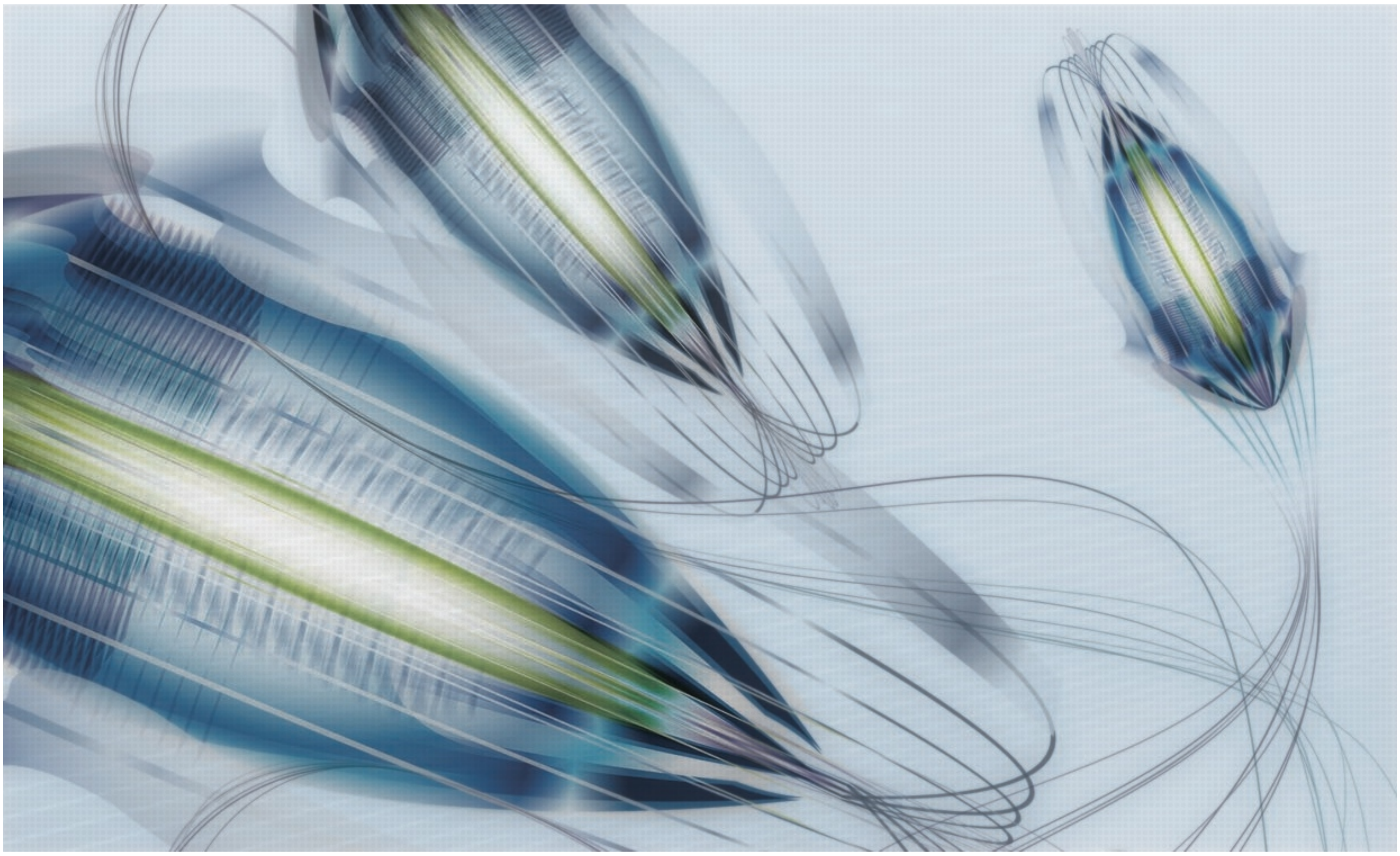

# Automatic Detection of Online Jihadist Hate Speech

## About the authors

**Tom De Smedt** is senior researcher at the Computational Linguistics & Psycholinguistics Research Center of the University of Antwerp (CLiPS), co-founder of the language technology company Textgain and co-founder of the Experimental Media Research Group at the St Lucas School of Arts Antwerp (EMRG). He has a PhD in Arts with a specialization in Computational Creativity, a Master's degree in Audiovisual Arts and a Bachelor's degree in software engineering.

**Guy De Pauw** has been working as a language engineer for over 20 years and has extensive experience developing text analytics applications for a wide range of languages. He has a PhD in Linguistics and is an associated researcher at the Computational Linguistics & Psycholinguistics Research Center of the University of Antwerp (CLiPS). He is co-founder and CEO of the CLiPS spin-off company Textgain, which develops Artificial Intelligence that performs author profiling in application areas such as online safety, security and human resources.

**Pieter Van Ostaeyen** has a Master's degree in Medieval History with a specialization in the history of the Crusades (KULeuven, 1999) and Arabic and Islamic Studies, focusing on the history of *Salah ad-Din al-Ayyubi and the Assassins* (KULeuven, 2003). He is an internationally recognized expert on the Islamic State and Belgian Foreign Terrorist Fighters, and among other he has published with CTC (Combating Terrorism Center, West Point, New York). He is a Visiting Fellow of the European Foundation for Democracy (EFD) and Member of the Editorial Board of the International Centre for Counter-terrorism The Hague (ICCT). He currently is pursuing his PhD at the department of Arabic and Islamic Studies at the University of Leuven, researching the usage of social media in the ideological strife between al-Qaeda and The Islamic State.

Prinsstraat 13, 2000 Antwerp, Belgium

+32 (0)3 265 52 22

# Automatic Detection of Online Jihadist Hate Speech

Tom De Smedt[1], Guy De Pauw[1], Pieter Van Ostaeyen[2]

[1] University of Antwerp, Computational Linguistics Research Group / Textgain

tom | guy@textgain.com

[2] University of Leuven, Arabic Studies

pieter.vanostaeyen@kuleuven.be

**Abstract.** We have developed a system that automatically detects online jihadist hate speech with over 80% accuracy, by using techniques from Natural Language Processing and Machine Learning. The system is trained on a corpus of 45,000 subversive Twitter messages collected from October 2014 to December 2016. We present a qualitative and quantitative analysis of the jihadist rhetoric in the corpus, examine the network of Twitter users, outline the technical procedure used to train the system, and discuss examples of use.

**Keywords:** jihadism, hate speech, text analytics, text profiling, machine learning

## 1 Introduction

Online hate speech is believed to play an important role in the advocacy of terrorism.[1] This is certainly true in the case of the Islamic State (الدولة الإسلامية, ISIS / ISIL / IS / Daesh), which has used an effective online propaganda machine to spread fundamentalist views on Islam, to recruit members and to incite fear and violence.[2] Examples include their digital magazine *Dabiq* and videos issued by the *al-Hayāt Media Center,*[3] but also the network of self-radicalized (or radicalizing) followers on social networks such as Twitter. On February 5, 2016, Twitter issued a press release to speak out against the use of their platform for the promotion of terrorism.[4] They reported suspending over 125,000 user profiles that endorsed acts of terrorism, remarking that "there is no 'magic algorithm' for identifying terrorist content on the internet, so global online platforms are forced to make challenging judgment calls based on very limited information and guidance."

---

[1] Robert S. Tanenbaum, "Preaching Terror: Free Speech or Wartime Incitement," *American University Law Review* 55 (2005): 785.

[2] Luis Tomé, "The 'Islamic State': Trajectory and reach a year after its self proclamation as a 'Caliphate'" (University of Lisbon, 2015).

[3] "Al-Hayat Media Center promotes barbaric execution video as leading top 10 list," *SITE Intelligence Group*, July 3, 2015, https://ent.siteintelgroup.com/Statements/al-hayat-media-center-promotes-barbaric-execution-video-as-leading-top-10-list.html

[4] "Combating Violent Extremism," *Twitter*, February 5, 2016, https://blog.twitter.com/official/en_us/a/2016/combating-violent-extremism.html

We have developed a system that can automatically detect jihadist hate speech with over 80% accuracy, using techniques from Natural Language Processing[5] and Machine Learning.[6] Such systems can help companies like Twitter to identify subversive content.

## HATE SPEECH

Hate speech can be defined as "any communication that disparages a person or a group on the basis of some characteristic such as race, color, ethnicity, gender, sexual orientation, nationality, religion, or other characteristic".[7] Hate speech is prohibited by law in several European countries such as Belgium, Denmark, France, Germany, Ireland, the Netherlands, Norway, Sweden and the United Kingdom. However, the European Court of Human Rights (ECHR) does not offer a legal definition of hate speech. Article 20 of the International Covenant on Civil and Political Rights (ICCPR) states that "any advocacy of national, racial or religious hatred that constitutes incitement to discrimination, hostility or violence shall be prohibited by law".[8] In the United States, hate speech is protected by the First Amendment and it has been argued that Article 20 of the ICCPR is unconstitutional according to the Supreme Court.[9] Nonetheless, as of May 23, 2016, several technology companies that play an unintended but significant role in the proliferation of hate speech, including Facebook, Google, Microsoft and Twitter, have jointly agreed to a European Union Code of Conduct to remove illegal online hate speech within 24 hours.[10]

## JIHADIST HATE SPEECH

Jihadist hate speech can be defined as hate speech proliferated by members or fans of Salafi jihadist militant groups, such as Al-Qaeda or ISIS, that is intended as propaganda, to incite violence or to threaten civilians, and which is either illegal (e.g., execution videos) or at least worrisome (e.g., promoting radicalization by framing nonbelievers as "worthless dogs").[11]

---

[5] Daniel Jurafsky and James H. Martin, *Speech and Language Processing* (Upper Saddle River: Prentice Hall, 2008).

[6] Fabrizio Sebastiani, "Machine Learning in Automated Text Categorization," *ACM Computing Surveys* 34, no. 1 (2002): 1-47.

[7] John T. Nockleby, "Hate speech," in *Encyclopedia of the American Constitution* (New York: Macmillan Publishers, 2000): 1277-1279.

[8] "International Covenant on Civil and Political Rights," *Office of the United Nations High Commissioner for Human Rights*, http://www.ohchr.org/EN/ProfessionalInterest/Pages/CCPR.aspx

[9] Michael Herz and Péter Molnár, eds., *The Content and Context of Hate Speech: Rethinking Regulation and Responses* (Cambridge University Press, 2012).

[10] Alex Hern, "Facebook, YouTube, Twitter and Microsoft sign EU hate speech code," *The Guardian*, May 31, 2016, https://www.theguardian.com/technology/2016/may/31/facebook-youtube-twitter-microsoft-eu-hate-speech-code

[11] Jytte Klausen et al., "The YouTube Jihadists: A Social Network Analysis of Al-Muhajiroun's Propaganda Campaign," *Perspectives on Terrorism* 6, no. 1 (2012).

ONLINE JIHADIST HATE SPEECH

Online jihadist hate speech can be found on social networks such as Twitter and Facebook, on video-sharing websites such as YouTube, on text-sharing websites such as JustPaste.it, and on encrypted chat apps such as Telegram, to name a few.[12] The content may include news updates, war photography, interviews with celebrity jihadists, fundamentalist interpretations of Islam, testimonial videos, execution videos, chants (*anasheed*), handbooks, infographics, advertisements, games and personal opinions, often expressed repetitively. In our experience, it is not hard to find if one knows what to look for.

## 2 Data collection

We have collected a corpus of online jihadist hate speech that consists of 49,311 "tweets" (public messages of no more than a 140 characters) posted by 367 Twitter users, henceforth the HATE corpus. As a counterweight, we have also collected a corpus of 35,166 tweets by 66 users that talk about Islam, Iraq, Syria, Western culture, and so on, without spreading hate speech, henceforth the SAFE corpus. Both corpora were extensively double-checked.

The SAFE corpus consists of tweets posted by reporters, imams and Muslims, for example `@HalaJaber` (a Lebanese-British journalist), `@TRACterrorism` (Terrorist Research & Analysis Consortium), `@BBCArabic`, `@TheNobleQuran` and `@AppleSUX` (an ISIS parody account). To balance the size of both corpora we added 15,000 random tweets from as many Twitter users to the SAFE corpus, on any topic from cooking to sports, in any language.

The HATE corpus was gradually expanded from October 2014 to December 2016. With each update we manually identified new subversive profiles and automatically collected tweets from all of them using the Pattern toolkit[13] and the Twitter API. Content was mainly collected in the aftermath of 10 incidents during this time period:[14]

1. **Charlie Hebdo shootings** (January 7, 2015). Saïd (34) and Chérif Kouachi (32) force their way into the offices of the French satirical newspaper Charlie Hebdo, armed with automatic rifles, killing 20 and injuring 22. We collected approximately 15,000 tweets.

[12] Ali Fisher. "How Jihadist Networks Maintain a Persistent Online Presence," *Perspectives on Terrorism* 9, no. 3 (2015).

[13] Tom De Smedt and Walter Daelemans, "Pattern for Python," *Journal of Machine Learning Research* 13 (2012): 2063-2067.

[14] "List of Islamist terrorist attacks," *Wikipedia*, November 24, 2017, https://en.wikipedia.org/wiki/List_of_Islamist_terrorist_attacks

2. **Sinai attacks** (January 29, 2015). Over 25 ISIS-affiliated militants target police offices, an army base and security checkpoints in Egypt, armed with mortars and car bombs, killing 44 and injuring 62. We collected approximately 10,000 tweets.
3. **Atatürk Airport attack** (June 20, 2015). Three or four ISIS-affiliated militants stage an attack at Turkey's international airport, armed with automatic rifles and explosive belts, killing 30 and injuring 104. We collected approximately 1,000 tweets.
4. **Paris attacks** (November 13, 2015). Three ISIS suicide bombers strike at Stade de France during a football match, followed by shootings at restaurants by another three and finally a shooting at the Bataclan theatre by three more, in total killing 137 and injuring 368. We collected approximately 7,500 tweets.
5. **Brussels bombings** (March 22, 2016). Five ISIS suicide bombers linked to the Paris attackers strike at Belgium's national airport and a Brussels metro station, killing 35 and injuring over 300. We collected approximately 5,000 tweets.
6. **Lahore bombing** (March 27, 2016). A suicide bomber targets Pakistan's largest park, killing 70 and injuring 300, most of them Muslim women and children. We collected approximately 3,000 tweets.
7. **Orlando nightclub shooting** (June 12, 2016). Lone wolf attacker Omar Mateen (29) enters a Florida LGBT nightclub, armed with a semi-automatic rifle, killing 49 and injuring 53, pledging allegiance to ISIS. We collected approximately 1,000 tweets.
8. **Nice attack** (July 14, 2016). Lone wolf ISIS supporter Mohamed Lahouaiej-Bouhlel (31) drives a truck into crowds of people celebrating France's Bastille Day on the Promenade des Anglais, killing 87 and injuring 434. We collected approximately 1,000 tweets.
9. **Botroseya Church bombing** (December 11, 2016). ISIS militant Mahmoud Shafiq Mohammed Mustafa (22) enters a renowned Coptic church in Egypt, armed with an explosive vest, killing 29 and injuring 47. We collected approximately 500 tweets.
10. **Berlin attack** (December 19, 2016). ISIS militant Anis Amri (24) hijacks a truck and drives it into a crowd at a German Christmas market, killing 12 and wounding 56. We collected approximately 1,500 tweets.

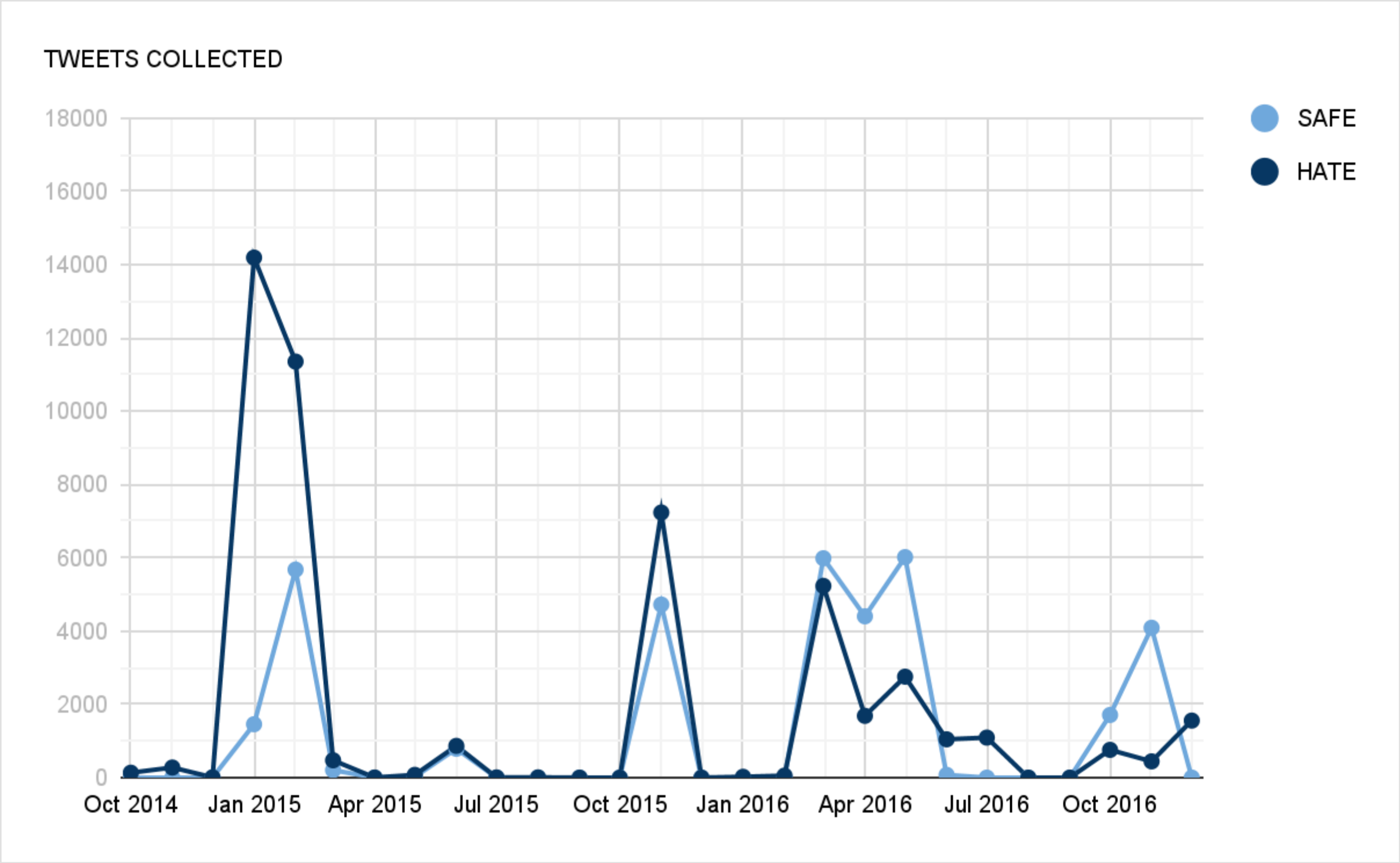

**Figure 1**. Timeline of tweets automatically collected from manually identified Twitter profiles.

All of the known perpetrators appear to be men and, where age is known, most of them young adults aged 20-35. Over time we observed how attacks became more improvised, with smaller teams of perpetrators often using more improvised weapons such as trucks. We also collected increasingly smaller amounts of tweets (see Figure 1). One explanation for the latter is that Twitter increasingly became more proactive in suspending subversive profiles and removing hate speech.[15] In 2017, jihadist hate speech became much harder to find on Twitter, as participants moved to encrypted, anonymous messaging apps such as Telegram.[16]

However, this does not mean that the problem has been solved, or that it may not resurface on public forums. While the move to Telegram can be seen as a tactical retreat, it may also have lacked the broad appeal of a public forum such as Twitter.[17] Today, new subversive Twitter profiles continue to appear and disappear in a relentless game of cat-and-mouse, as organizations such as ISIS continue to strive to spread their views.[18] Consequently, a report issued by the European Commission on September 28, 2017 has outlined new guidelines on illegal hate speech and terrorism-related content, urging Facebook, Google, Microsoft and Twitter to improve their detection capabilities.[19]

---

[15] Adam Satariano, "Twitter Suspends 300,000 Accounts Tied to Terrorism in 2017," *Bloomberg*, September 19, 2017, https://www.bloomberg.com/news/articles/2017-09-19/twitter-suspends-300-000-accounts-in-2017-for-terrorism-content

[16] J. M. Berger and Heather Perez, "The Islamic State's Diminishing Returns on Twitter: How suspensions are limiting the social networks of English-speaking ISIS supporters" (George Washington University, 2016).

[17] Gabriel Weimann, "Terrorist Migration to the Dark Web," *Perspectives on Terrorism* 10, no. 3 (2016).

[18] "IS taaier online dan in het veld," *De Morgen*, September 22, 2017, https://www.demorgen.be/plus/is-taaier-online-dan-in-het-veld-b-1506037201339/

## 2.1 Manual identification

For data collection, we first identified 367 subversive profiles, by regularly searching Twitter for keywords such as *kuffar* (كفّار, unbelievers) and examining profiles that use these words in their tweets. The combination of a profile's username, profile picture, names of places and rhetoric can be used to reliably identify subversive profiles. Once a few profiles have been found, it is not hard to find more by examining their network of friends, including profiles practicing countermeasures against identification (e.g., choosing `@1Ak187` as a username).

**By username.** One useful cue to identify a target profile is its username. Each Twitter account has a unique alphanumeric username that appears with every message and that can be used to mention or cite (retweet) another user by prepending the @ symbol. For example:

> RT @pioiuhghsd42424: #Bruxelles Après les bombes, attendez nos soldats en kalachnikovs. Ils vont mitrailler dans toutes les rues !ي
>
> Posted by `@nightwalker_118` (March 22, 2016)

Twitter users tend to choose a username that reflects their actual name, gender, personality or occupation.[20] ISIS and similar groups have standardized conventions for *noms de guerre* (كنية, *kunya*), which start with *Abu* and end with the tribe or place of origin, as in Abu Bakr al-Baghdadi or Abu Hamza al-Muhajir, where *Muhajir* denotes a Muslim immigrant, more specifically a foreign Jihad fighter.[21] In combination with other cues it is useful to scan the username of a profile for *abu*, *mujahid*, *muhajir*, and so on, as illustrated in Figure 2. In this example, the username (`@AbuluqmanIS`) in combination with the actual name (Abu Luqman al-Muhajir) and the profile picture is a strong cue. Reading the user's tweets then confirms it.

---

[19] "Communication on Tackling Illegal Content Online - Towards an enhanced responsibility of online platforms," *European Commission*, September 28, 2017, https://ec.europa.eu/digital-single-market/en/news/communication-tackling-illegal-content-online-towards-enhanced-responsibility-online-platforms

[20] Jing Liu et al., "What's in a Name? An Unsupervised Approach to Link Users Across Communities," in *Proceedings of the Sixth ACM International Conference on Web Search and Data Mining*, 495-504, 2013.

[21] Vera Mironova and Karam Alhamad, "The Names of Jihad: A Guide to ISIS' Noms de Guerre," *Foreign Affairs*, July 14, 2017, https://www.foreignaffairs.com/articles/syria/2017-07-14/names-jihad

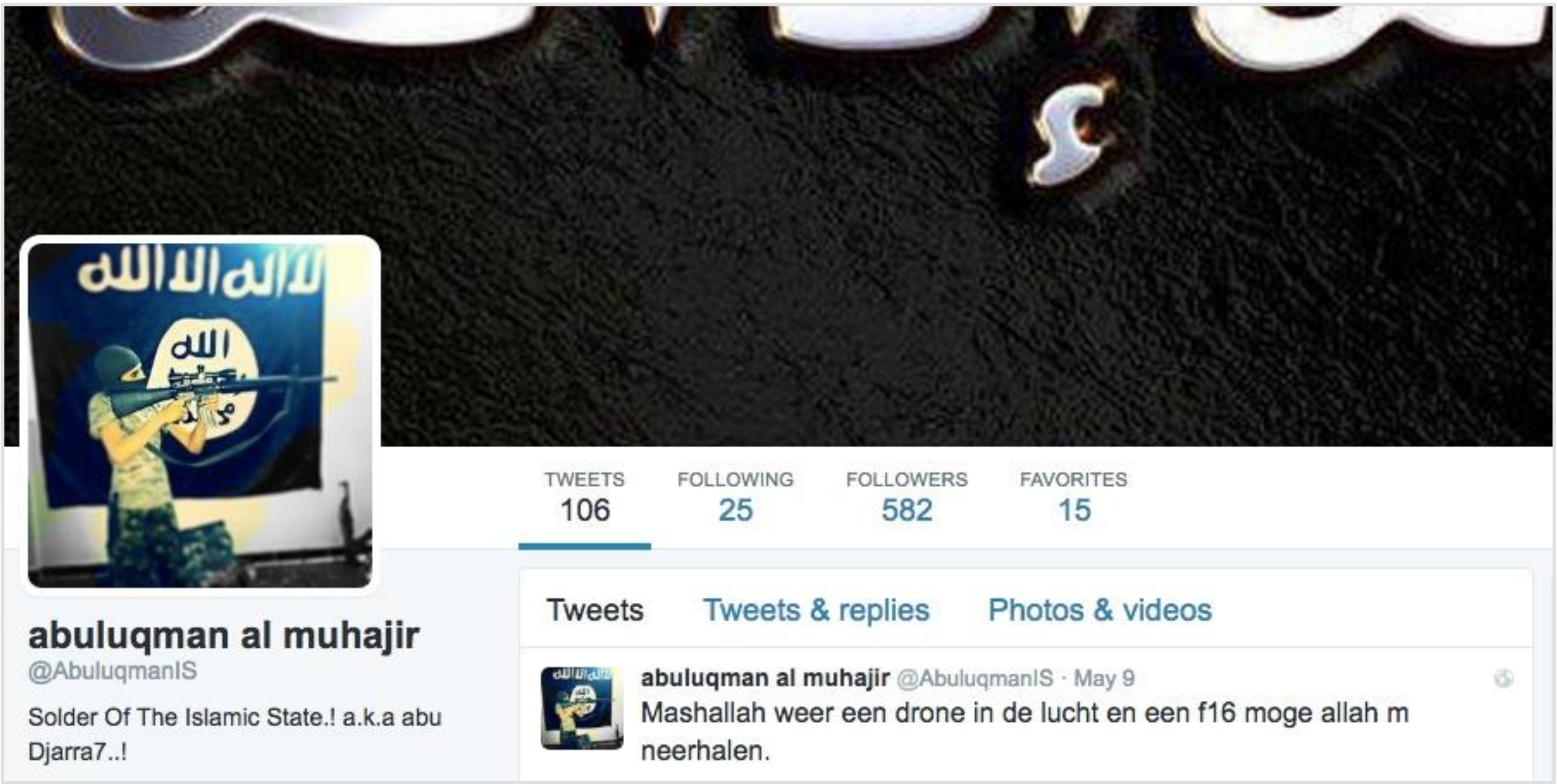

**Figure 2**. Profile identification by username.

In our dataset of 367 Twitter profiles, we have 40 usernames that start with *Abu* (father, e.g., `@AbuHamzaIS`), we have 6 that start with *Umm* (mother, e.g., `@_UmmWaqqas`), 5 that contain *Muhajir* (`@Muhajir_Miski1`) and 4 others that contain *Jihad* (`@JihadiA6`). Unfortunately, nowadays usernames of subversive profiles are mostly random, e.g., `@pioiuhghsd42424` or `@c0n0fj1had4_`, no doubt in an attempt to hide the user's identity. Monitoring users as they move from one anonymous account to the next anonymous account is challenging because we only have their writing style and retweets as a given.

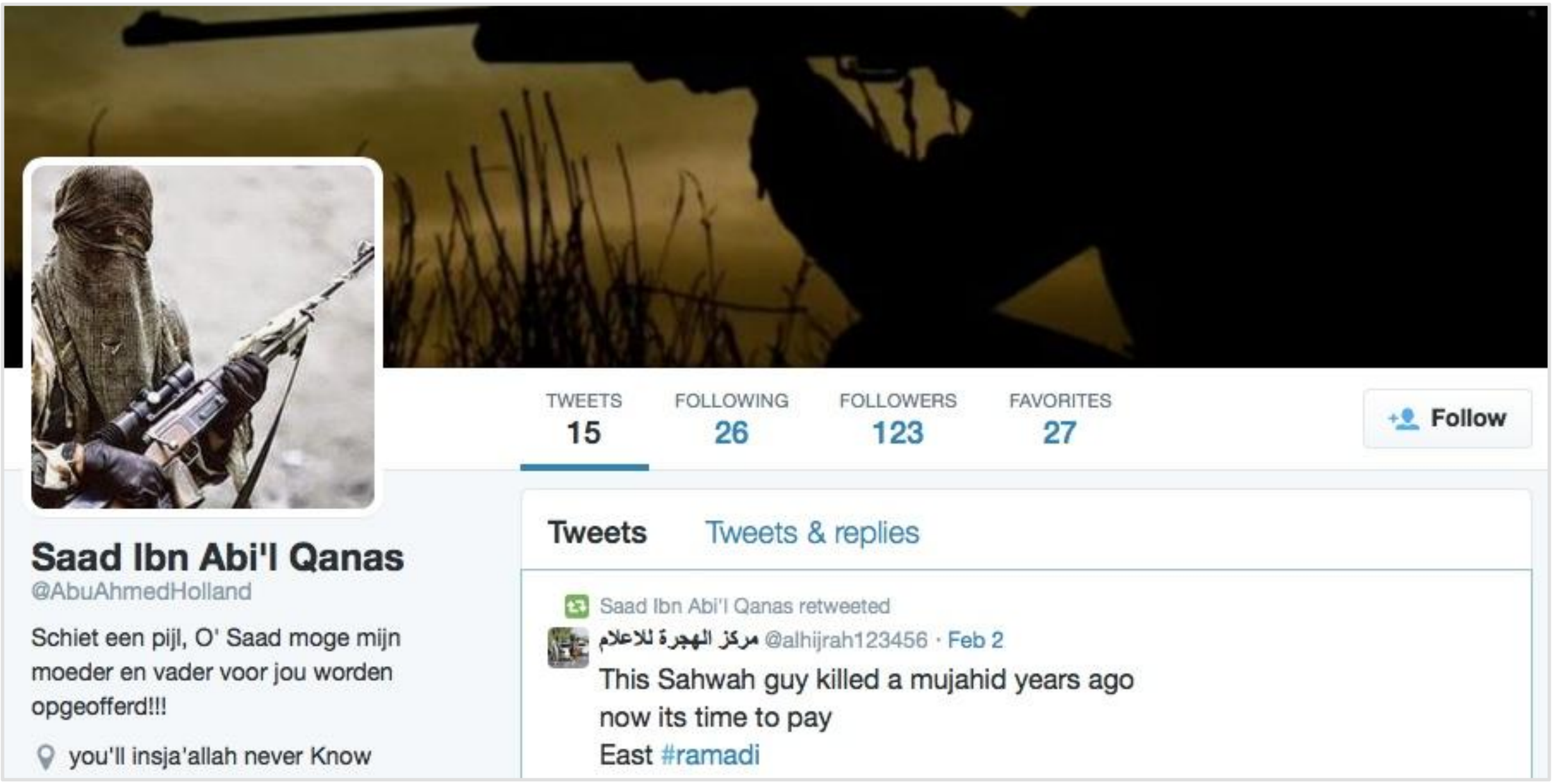

**Figure 3**. Profile identification by picture.

**By picture.** Another useful cue is the profile picture, often showing a masked person holding a firearm, or a landmark such as the Eiffel Tower in flames, or the ISIS flag. Interestingly, some profile pictures display a lion, cat or kitten, which presumably serves as a metaphor for bravery and martyrdom. Figure 3 shows the profile of a masked person with a scoped rifle. In combination with the profile's username (`@AbuAhmedHolland`) and a tweet that mentions `#ramadi` (i.e., a city in Iraq on the route to Syria) it is a strong cue.

**By location.** Another useful cue is the user's location. Some Twitter profiles broadcast their geolocation coordinates in their metadata[22] while other profiles may mention names of places in war zones and/or post photos of the local area.

**By rhetoric.** Jihadist hate speech is grounded in a selective and fundamentalist (Salafist) view on the Quran[23] that can be difficult to spot by outsiders, particularly because of spelling variations used by different foreign fighters. But some recurring words for which even Google Translate and Wikipedia offer definitions stand out. For example, *takfīr* (تكفير) refers to the excommunication of infidel Muslims, perhaps because of *nifaq* (نفاق), hypocrisy and insincerity with grave consequences in the afterlife, or *shirk* (شرك), the sin of worshipping gods besides Allah. Using such words in Twitter Search yields interesting hits (Figure 4).

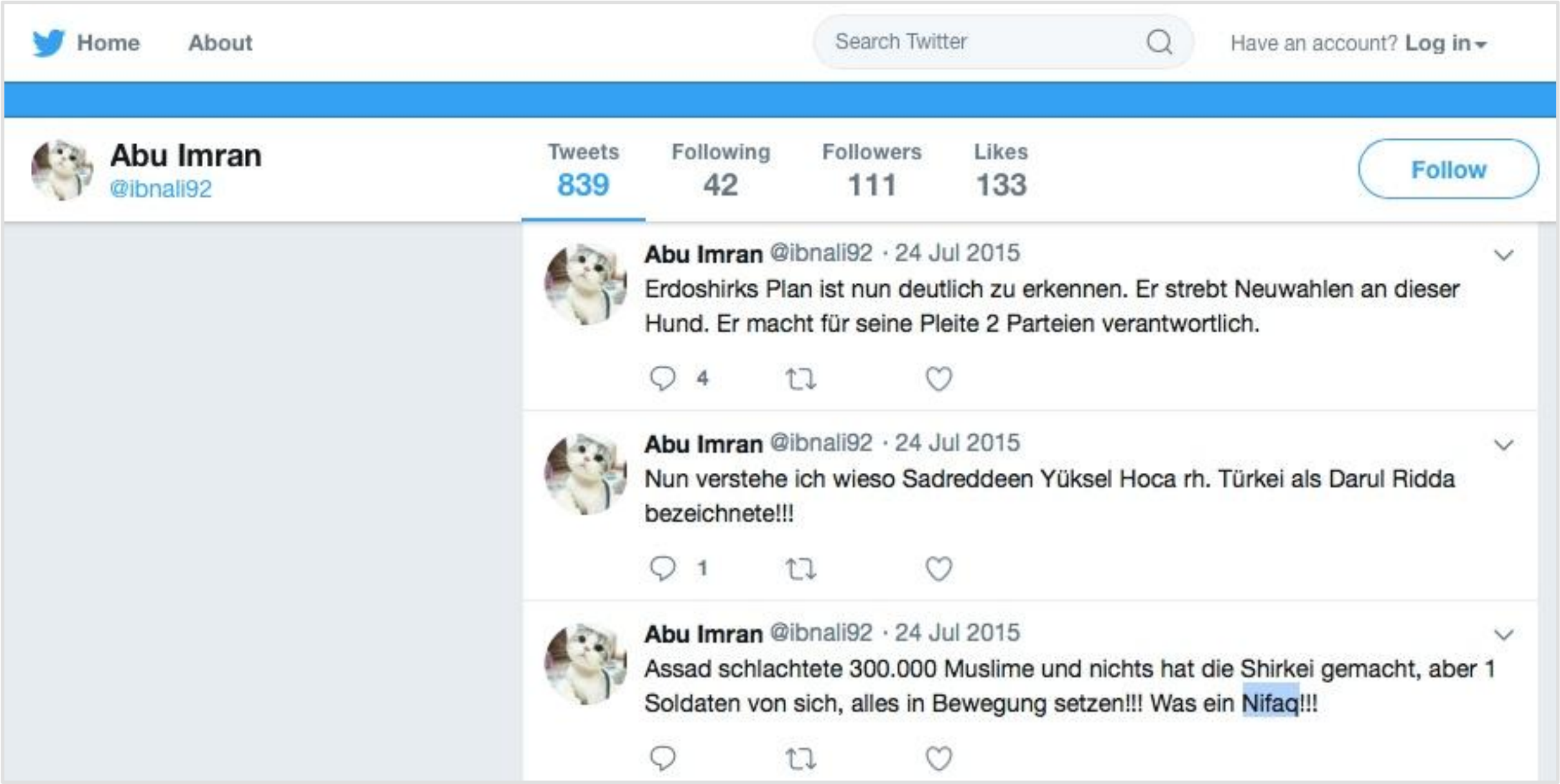

**Figure 4**. Profile identification by language use.

[22] Elizabeth Bodine-Baron et al, "Examining ISIS Support and Opposition Networks on Twitter" (RAND, 2016).

[23] Donald Holbrook, "Using the Qur'an to Justify Terrorist Violence: Analysing Selective Application of the Qur'an in English-Language Militant Islamist Discourse," *Perspectives on Terrorism* 4, no. 3 (2010).

## 2.2 Automatic collection

We used the Pattern toolkit for the Python programming language to automatically collect tweets posted by our set of manually identified profiles. A sample is shown in Figure 5. An example Python script is shown below. It will store the most recent tweets posted by `@BBC` in a CSV file (comma-separated values) while discarding duplicates that it has already seen.

| | A | B | C |
|---|---|---|---|
| 1 | 556392744936284160 | Muhajir_Miski1 | RT @AHudhayfah: To the hypocrites/extreme modernists who defend the people who insult the #ProphetMuhamm |
| 2 | 556386119957614592 | Muhajir_Miski1 | @Umm_Umayr Wa Iyyaki ukht |
| 3 | 556383309086064640 | Muhajir_Miski1 | @Umm_Umayr When he comes back insha Allah |
| 4 | 556370123268059136 | Muhajir_Miski1 | RT @Protocol_Shami2: Islam has become Pasicfism, our way of Life reduced to Subguation, Men have turned int |
| 5 | 556369538896654337 | Muhajir_Miski1 | RT @Najween__: Algerian Muslims protesting against the slandering of our beloved Prophet, shouting: "We want |
| 6 | 556369219785605120 | Muhajir_Miski1 | @LittleMenus We have no love for Homosexuals @ShariahisComing |
| 7 | 556369036188332032 | Muhajir_Miski1 | @GTvehicle I take that as a compliment @hudaifyhudaify |
| 8 | 556344472590884864 | Muhajir_Miski1 | @bani_diran No problem akhil fadil |
| 9 | 556343899682521088 | Muhajir_Miski1 | @bani_diran Akhi I'm not cursing them specificly. I'm cursing the Homo's in general but Barakallahu feek |
| 10 | 556343337243140096 | Muhajir_Miski1 | @ShariahisComing I know akhi, May Allah swt reward the ones who've made this happen |
| 11 | 556342494288699392 | Muhajir_Miski1 | A good homosexual is a dead one. Alhamdulillah the Mujahideen have implemented the Rule of Allah on them. #S |
| 12 | 556341201662930945 | Muhajir_Miski1 | This is the punishment for the Homosexuals. May the curse of Allah be upon them. http://t.co/4WWJd85PWb |
| 13 | 556042280155766784 | Muhajir_Miski1 | @hudaifyhudaify Islam does not only mean peace, It also means War. |
| 14 | 556042170562789378 | Muhajir_Miski1 | @hudaifyhudaify If u dont believe me, Go ask a knowledgeable sheikh what the punishment for the one who insul |
| 15 | 556040996887470081 | Muhajir_Miski1 | @hudaifyhudaify Alhamdulillah I left my home to implement it. Starting from myself and the rest of the Ummah. |

**Figure 5**. Automatic collection of tweets into a CSV file (displayed in Google Sheets).

```
# coding: utf-8
# http://github.com/clips/pattern

from pattern.db  import Datasheet
from pattern.web import Twitter

try:
    data = Datasheet.load('tweets.csv')
    seen = data.columns[0]
except:
    data = Datasheet()
    seen = []

twitter = Twitter()

for username in ('@BBC',):
    for tweet in twitter.search('from:' + username, count=100):
        if tweet.id not in seen:
            data.append((
                tweet.id,
                tweet.author,
                tweet.text,
                tweet.date
            ))

data.save('tweets.csv')
```

# 3 Data evaluation (HATE)

## 3.1 Qualitative analysis

The HATE corpus primarily consists of tweets posted in direct response to terrorist attacks. We did not study how tweets prior to an attack might have fueled the event. Following is a selection of tweets in the corpus and a short discussion of their characteristics. In general, jihadist rhetoric on social media appears to be riddled with spelling errors, slang, sarcasm, juvenile boasting, bonding (e.g., أخ, akh, brother) and emoticons – in particular the raised index finger (☝) and tears of joy (😂).

> Europe will spend the rest if its miserable days on tender hooks ☝
> Posted by `@bintislamiya19` (March 22, 2016)

> Les mecs executent les otages trkl oklm, malgre le fait que l'armee (aussi incompetente soit-elle) les traque 😂 #fusillade
> Posted by `@Aljabarti45` (November 13, 2015)

> Allahu Akbaaaar Allahu Akbaaaaar Daulatul Islamiya baqiya watatamadad
> DIE IN YOUR RAGE #KUFFARES
> Posted by `@AlTaifatul11` (November 14, 2015)

> Did the kuffar think tht they will be left alone after harming us?
> Have a taste of ur own medicine #brusselsairport
> Posted by `@Jundullah40` (March 22, 2016)

In the first example, the raised index finger refers to Tawhid (توحيد), the fundamental concept in Islam that God is one and absolute. More specifically, Salafi jihadist organizations such as ISIS adhere to a fundamentalist interpretation of Tawhid that rejects non-fundamentalist regimes as idolatrous, demanding destruction.[24]

In the second example, the author is mocking the police during the Bataclan raid, where the perpetrators (*les mecs*, the guys) executed their hostages. Urbandictionary.com defines *trkl* as a French phonetic deformation of *tranquille* (be calm) and *oklm* as *au calme* (be cool).

---

[24] Nathaniel Zelinsky, “ISIS Sends a Message: What Gestures Say About Today’s Middle East,” *Foreign Affairs*, September 3, 2014, https://www.foreignaffairs.com/articles/middle-east/2014-09-03/isis-sends-message

In the third and fourth examples, threats are made against the "unbelievers" (kuffar, kufr, dogs, pigs), a prevalent theme in jihadist rhetoric.[25] Tweets may be interspersed with Arabic words or phonetic transliterations, such as *dawla* or *daulat* for al-Dawla (الدولة, the state). In the third example, *Daulatul Islamiya baqiya watatamadad* refers to ad-Dawlah al-Islāmiyah (the Islamic State) and its slogan "Baqiya wa Tatamadad", Lasting and Expanding.[26] The colorful writing style in jihadist rhetoric is advantageous, because it is more predictable.

The HATE corpus includes some well-known ISIS supporters, at least in hindsight. Below is an overview of those that we were able to identify using publicly available resources:

- **Abu al-Baraa el-Azdi** (`@AbuAlbaraaSham`). Saudi preacher that became the religious judge of the city of Derna in Libya, which he declared to be a franchise of the Islamic State in October 2014.[27]
- **Anjem Choudary** (`@anjemchoudary`). British spokesman of Islam4UK, an activist group in the United Kingdom, who was sentenced to five years and six months in prison on September 6, 2016 on terrorism charges.
- **Junaid Hussain** (`@abuhussain1337_`). British hacker of the Cyber Caliphate and high-profile propagandist of lone wolf terrorism, who was killed in Syria on August 24, 2015 by a targeted drone strike.[28]
- **Rawdah Abdisalaam** (`@_UmmWaqqas`). Female student of journalism based in Seattle[29] (possibly of Finnish or Dutch origin) who became a major online ISIS recruiter, and is believed to have traveled to Syria.

Other known profiles include Sally Jones (`@UmmHussain107`), the wife of Junaid Hussain, Anis Abou Bram (`@abubrams`), a radicalized Belgian teen killed in Syria in February 2015, and Abu Abdullah Britani (`@abu_britani2`), who was ridiculed online for threatening to take over Rome and throw homosexuals off "your leaning tower of pizza" (sic).

---

[25] Anina L. Kinzel, "From Keywords to Discursive Legitimation: Representing 'kuffar' in the Jihadist Propaganda Magazines," in *Proceedings of Corpus Linguistics Fest 2016*, 26-33, 2016.

[26] Omar Ashour, "Enigma of 'Baqiya wa Tatamadad': The Islamic State Organization's Military Survival" (Al Jazeera, 2016).

[27] Paul Cruickshank et al., "ISIS comes to Libya," *CNN*, November 18, 2014.

[28] Stefano Mele, "Terrorism and the Internet: Finding a Profile of the Islamic 'Cyber Terrorist'," in *Countering Terrorism, Preventing Radicalization and Protecting Cultural Heritage: The Role of Human Factors and Technology* (Amsterdam: IOS Press, 2017): 103.

[29] "Rawdah Abdisalaam a.k.a. @_UmmWaqqas," *Counter Extremism Project*, https://www.counterextremism.com/extremists/rawdah-abdisalaam-aka-ummwaqqas

## 3.2 Quantitative analysis

TEXT ANALYTICS

Over the past decade, tools for Natural Language Processing (NLP) have become faster and more reliable, to the extent that they can now be applied to analyze large quantities of text in real-time (i.e., Big Data). For example, part-of-speech tagging can be used to detect word types (e.g., noun, verb) based on a word's position in the sentence, topic detection can be used to categorize articles (politics, sports, etc.), and sentiment analysis can be used to detect the tone of voice (e.g., positive, negative) in product reviews.[30] This is called text analytics.

TEXT PROFILING

More recent systems can also be used to detect the age, gender, education or personality of authors.[31] This is called text profiling. Text profiling relies on the writing style of different individuals.[32] For example, statistically, women tend to use more personal pronouns (*I*, *you*, *we*) to talk about people and relationships, while men use more determiners (*a*, *the*) and quantifiers (*one*, *many*) to talk about objects and concepts. Adolescents use more informal language such as abbreviated utterances (*omg*, *wow*) and adjectives (*awesome*, *lame*) to talk about school, parents and partying, while adults use more complex sentence structures and less emoticons to talk about work, children and health, and so on.

We used the Textgain API[33] to automatically detect the language and tone of voice of each tweet in the HATE corpus, the main keywords and names of places, and the age, gender and education level of the authors. The Textgain API operates by analyzing a text that is sent to the secure server and by responding with, for example, `M` or `F` for gender, or `25+` or `25-` for age. The aggregated results are discussed below and represented in Figure 7.

---

[30] Tom De Smedt and Walter Daelemans. "'Vreselijk mooi!' (terribly beautiful): A Subjectivity Lexicon for Dutch Adjectives," In *Proceedings of the 8th Language Resources and Evaluation Conference*, 3568-3572, 2012.

[31] Ben Verhoeven, Walter Daelemans, and Tom De Smedt, "Ensemble Methods for Personality Recognition," in *Proceedings of the Workshop on Computational Personality Recognition*, 35-38, 2013.

[32] James W. Pennebaker, *The Secret Life of Pronouns: What Our Words Say About Us* (New York: Bloomsbury Press, 2011).

[33] "Textgain: Web services for predictive text analytics," *Textgain*, https://www.textgain.com

**Language.** The Textgain API has an accuracy of 95% for language detection, meaning an error rate of 1/20. The HATE corpus mainly consists of tweets written in English (about 40%) and Arabic (30%). About 20% of tweets are written in French, Italian, Portuguese, Swedish, Dutch, Farsi, Turkish and German. All other languages make up about 10% of the corpus, mainly Finnish, Indonesian, Latvian, Danish and Spanish.

**Region.** We scanned each tweet for mentions of a country's name or capital city. Syria is mentioned in 1,191 tweets (about 2.5%), Iraq in 932 (about 2%) and France in 667 (1.5%). Other countries mentioned include the United States, Israel, Russia, Jordan, Iran, Egypt and Yemen. Paris is mentioned in 921 tweets, Baghdad in 217 and Brussels in 179. Other capital cities mentioned include Doha, Damascus, London and Jerusalem.

**Age & gender.** We used the API to detect the age and gender of each author. We only used English tweets (i.e., 40% of the corpus) for which the API has an accuracy of 75%. We grouped tweets by author, since more text gives more reliable results. About 95% of authors are predicted to be adults (25+) and 5% adolescents (25-). About 95% are predicted to be men (M) and 5% women (F).

**Tone.** We used the API to detect the tone of voice (positive, negative or neutral) in each tweet, for which the API has an accuracy of 75%. About 35% of the tweets are predicted as positive, 45% as negative and 20% as neutral. By comparison, the SAFE corpus has 50% neutral tweets and only 15% negative tweets. This is because it consists of news

| LANGUAGE | | TWEETS | % |
|---|---|---|---|
| EN | English | 19063 | 38.7% |
| AR | Arabic | 13675 | 27.7% |
| FR | French | 2304 | 4.7% |
| IT | Italian | 1762 | 3.6% |
| PT | Portuguese | 1747 | 3.5% |
| SV | Swedish | 1691 | 3.4% |
| NL | Dutch | 1684 | 3.4% |
| FA | Farsi | 1101 | 2.2% |
| TR | Turkish | 986 | 2.0% |
| DE | German | 897 | 1.8% |
| | ... | 4401 | 8.9% |

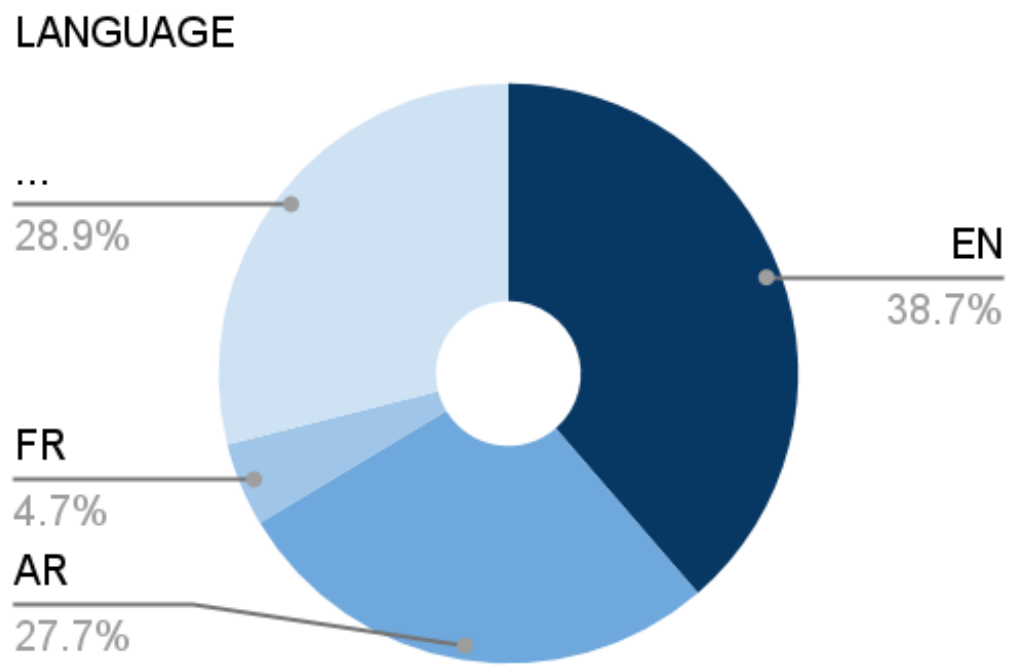

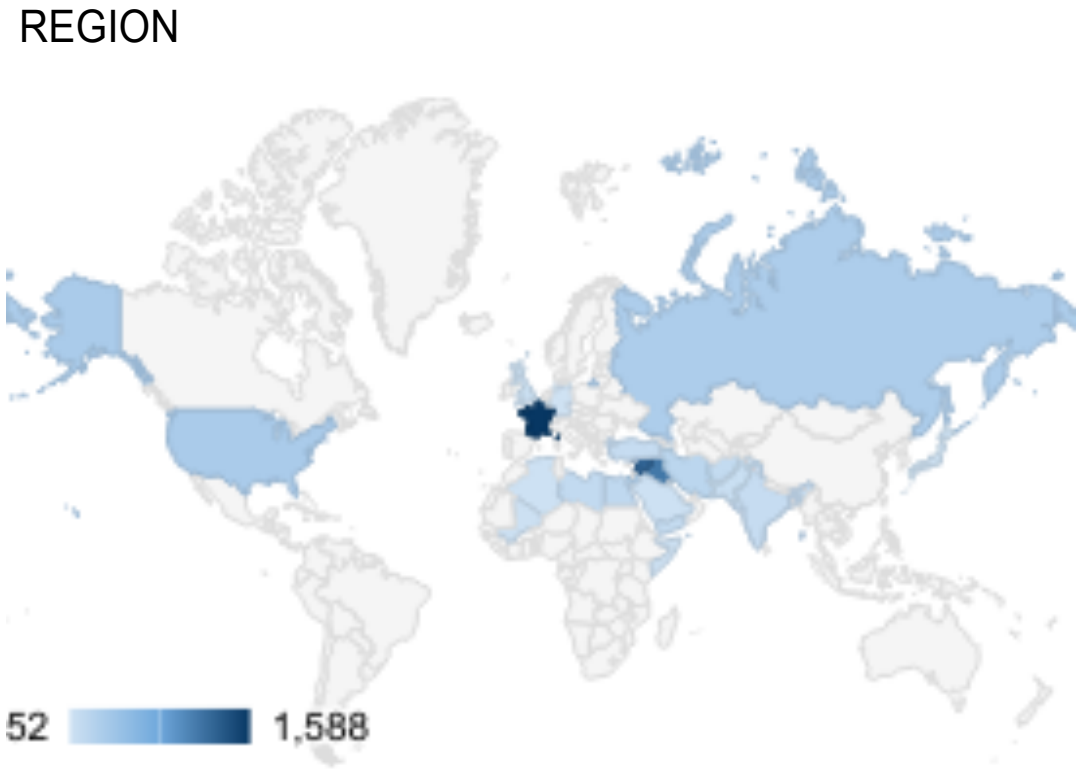

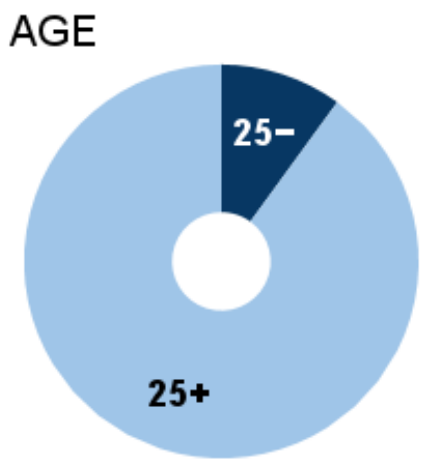

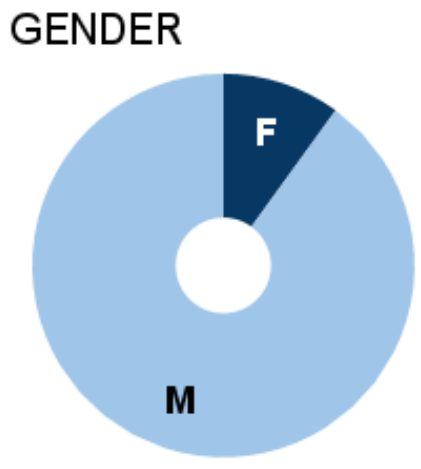

updates and tweets by ordinary people that do not talk about stupid kuffar, evil cretins, filthy dogs, vile French crusaders, and so on.

**Education.** About 60% of authors are predicted as educated (fluent writing) and 40% as less educated (poor writing / foreign speaker). Accuracy: 80%.

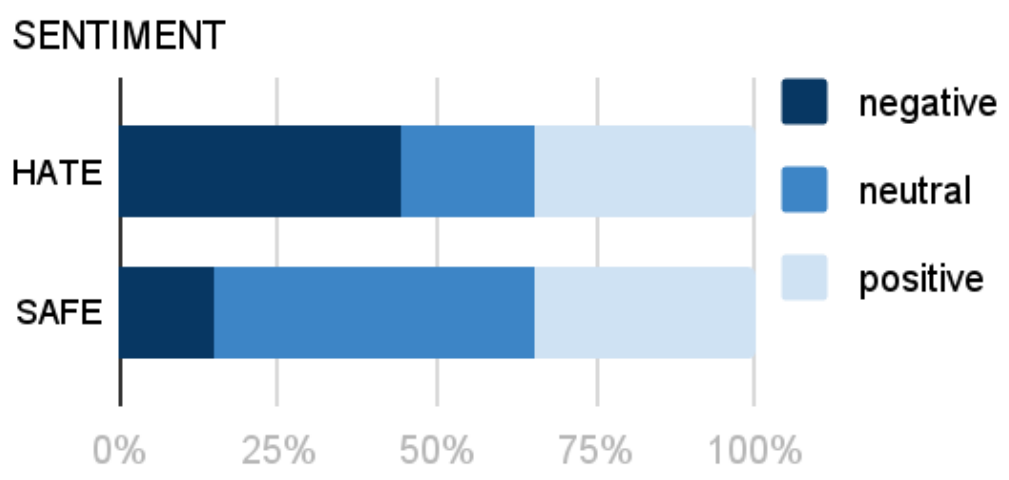

**Figure 7**. The HATE corpus in charts.

## 3.3 Keyword analysis

What is the probability that a given word occurs in a hateful tweet? While function words such as *the* and *we* will occur in any kind of tweet, the rationale is that content words such as *football* or *jihad* will occur more often in specific kinds of tweets (i.e., sports tweets vs. hate speech). Conversely, their occurrence can serve as a good cue for determining what kind of tweet we are dealing with. To assess which words are significantly biased, we used a chi-squared test[34] with $p \leq 0.01$. As it turns out, thousands of words are significantly biased. The chi-squared test does not tell us to what kind of tweet (HATE / SAFE) but we can also calculate the posterior probability $P(\theta|x)$ that a word occurs in a hateful tweet. This exposes keywords such as *crusader*, *curse*, *khilafah*, *kuffar*, *martyrdom*, *rage*, *shirk*, and so on. Independently, these words do not necessarily need to raise a red flag, but rather a **combination** of them does, as in: "Die in your rage kuffar!" (Figure 8).

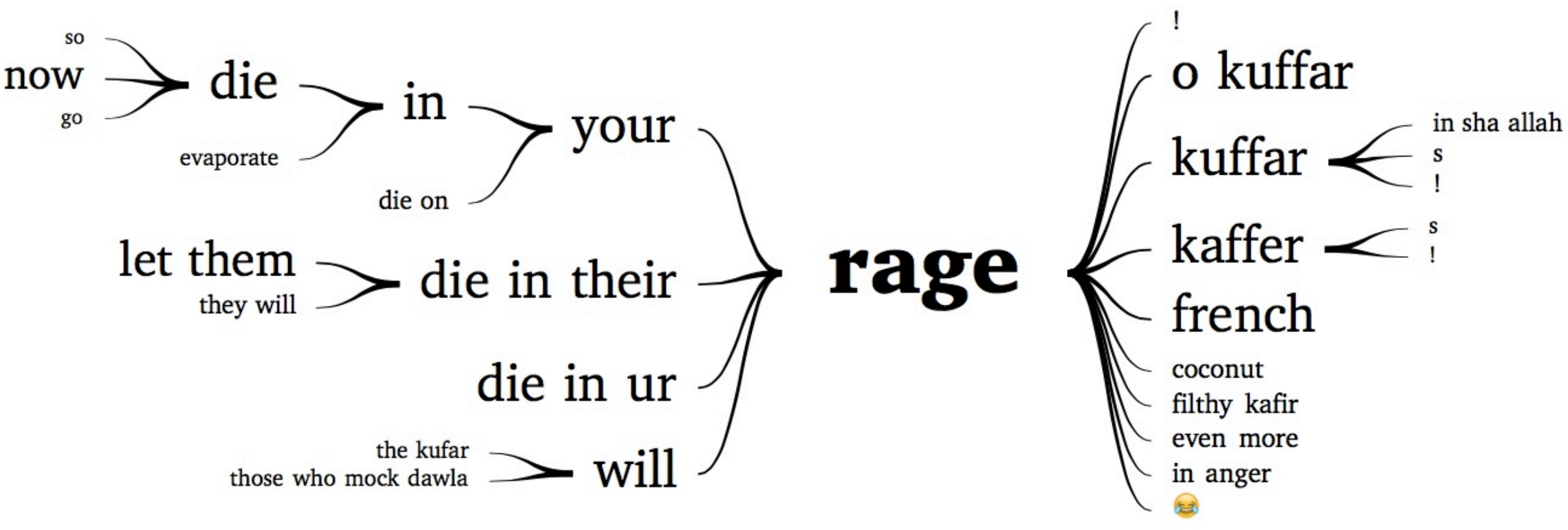

**Figure 8**. Word tree showing the context for "rage" in the HATE corpus.

[34] Huan Liu, and Hiroshi Motoda, eds., *Computational Methods of Feature Selection* (Chapman and Hall/CRC Press, 2007).

Figure 9 below provides a random sample of biased words, each with a short description, the probability that a tweet constitutes hate speech if the word appears in it (%) and the total number of times (#) that the word appears in the HATE + SAFE corpora:

| KEYWORD | DESCRIPTION | HATE % | # |
|---|---|---|---|
| jihad | literally: struggling; ideologically: war against unbelievers | 79% | 978 |
| kafir, kuffar, kufr | unbeliever(s) | 98% | 1349 |
| shirk | idolatry = worshipping other gods but Allah | 96% | 111 |
| mujahid | jihad fighter | 96% | 776 |
| muhajir | jihad fighter from a foreign country | 80% | 359 |
| lion | jihad fighter with bravery | 81% | 423 |
| murtad | a Muslim that is an apostate (i.e., rejects Islam) | 95% | 307 |
| takfiri | a Muslim that accuses another Muslim of apostasy | 86% | 123 |
| sharia | roughly: religious law | 88% | 305 |
| dawla | roughly: state | 66% | 840 |
| umma | roughly: the community of all Muslims across states | 90% | 737 |
| khilafah, الخلافة | the caliphate | 83% | 543 |
| coconut | kafir (i.e., brown outside but white inside) | 86% | 177 |
| dogs, pigs | US or European Christians, Shia Muslims, Jews, atheists, … | 75% | 229 |
| kill | - | 67% | 2560 |

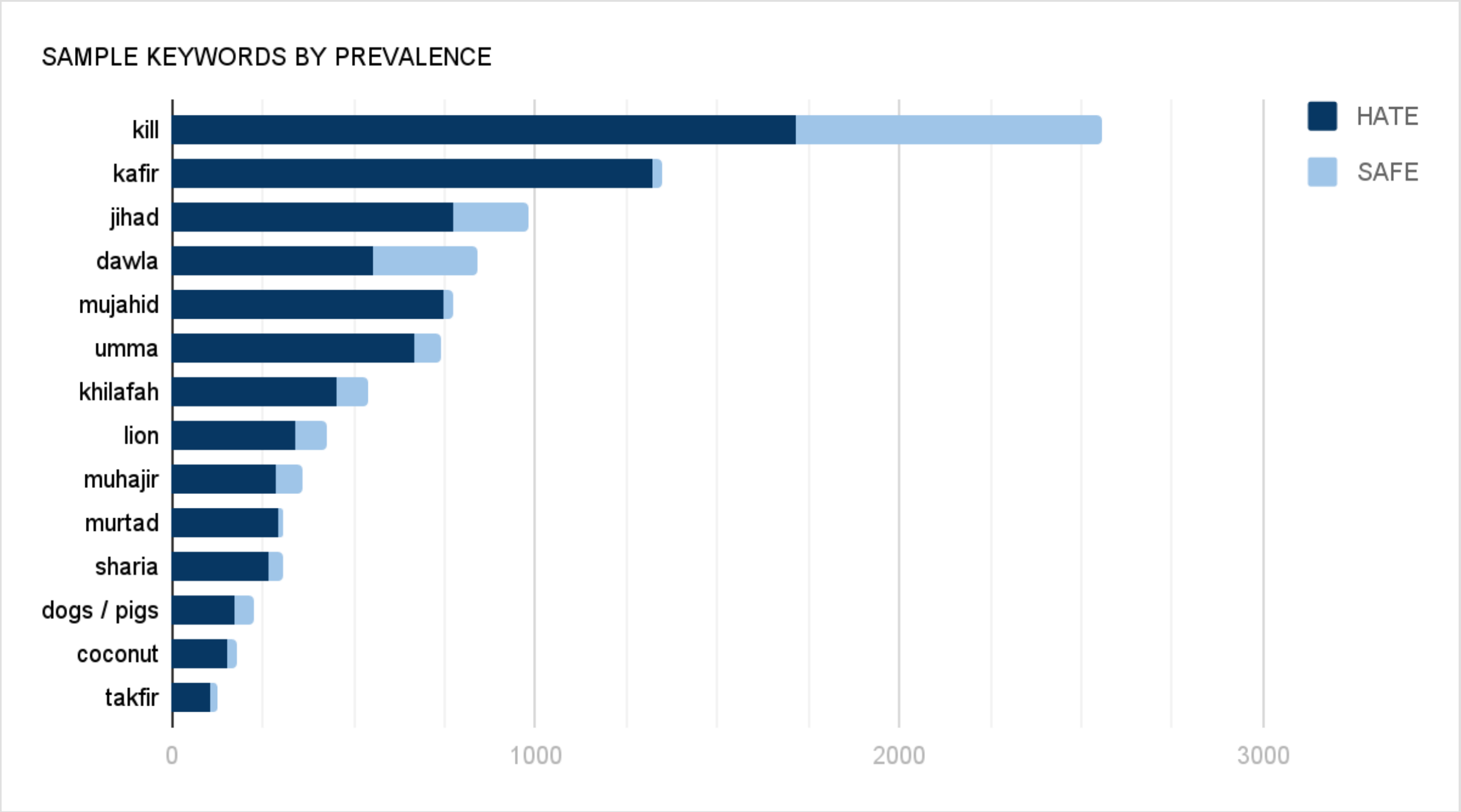

**Figure 9**. Sample keywords in jihadist rhetoric.

Figure 10 shows a word cloud with a sample of 200 significantly biased words ($p \leq 0.01$) that occur very frequently in hate tweets ($P(\theta|x) > 95\%$):

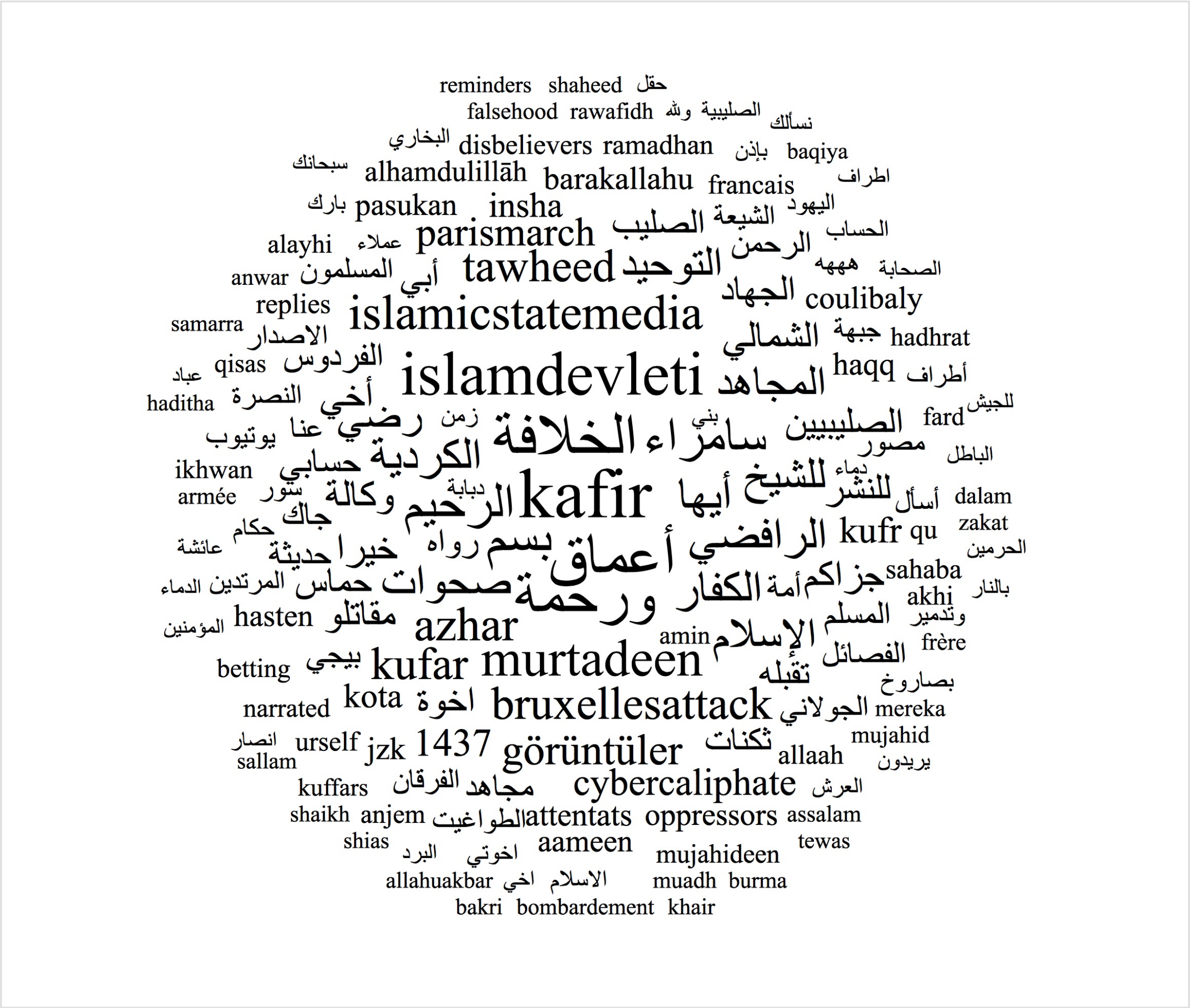

| ARABIC | ENGLISH | DESCRIPTION |
| --- | --- | --- |
| بسم | in the name | in the name of Allah the Merciful, بسم الله الرحمن الرحيم |
| الخلافة | the caliphate | Islamic State of Iraq and Syria (ISIS) |
| أعماق | amaq (news agency) | news outlet that claims responsibility for ISIS |
| رحمة | mercy | - |
| سامراء | Samarra | holy city in Iraq; possibly the birthplace of Abu Bakr al-Baghdadi |
| المجاهد | the mujahid | jihad fighter |
| الكفار | the infidels | persons with no religion |
| الرافضي | al-rafidi | rejectionists (esp. Shia Muslims) |
| التوحيد | tawhid | The Oneness of God |

**Figure 10**. Sample keywords in jihadist rhetoric.

In the case of emoticons, the raised index finger (☝) has a significant bias (96%) as does the tears of joy (😂) and the arrow pointing down (⬇) – in combination with other keywords of course. The arrow down is used to draw attention to new (previously suspended) profiles.

On a general level, the rhetoric in online jihadist hate speech is religiously polarizing, with vitriolic references to Western unbelievers, apostates, crusaders, and so on. This is in line with the proposition that the jihadist narrative attempts to frame an Islam that is under attack by the West, and that all Muslims must unite against it.[35] The global unity is implied by the *umma*, to which the *dawla* (state) is a means to an end.[36]

On a fine-grained level, the keyword analysis exposes unlikely derogatory metaphors, such as *coconut,* which denotes a so-called moderate: brown (Muslim) on the outside, white (kafir) on the inside. New instances of such inside language use would be hard to uncover by hand.

## 4 Automatic prediction

Machine Learning (ML) is a field related to Artificial Intelligence (AI) that uses statistical approaches to "learn by example". For example, when given 10,000 English texts and 10,000 French texts, a machine learning algorithm will automatically discover prevalent linguistic patterns that can then be used to predict whether another text is written in English or in French (e.g., word endings such as *-ized* are good cues for English while diacritics such as é and ç are good cues for French).

Machine learning algorithms expect their learning examples to be given as a set of *vectors*, where each vector is a set of feature → weight pairs[37]. For text classification, the features could be words, and the weights could be word count. We use character trigrams as features. Character trigrams are sequences of three successive characters, for example *kuffar* = { `kuf`, `uff`, `ffa`, `far` }. The advantage of this approach is that it efficiently models word endings, function words, emoticons, as well as spelling variations (e.g., the vectors for *kuffar* and *kufr* have a match on `kuf`). To prevent overfitting, we anonymized URLs and usernames (e.g., `@StaatsNieuws` becomes `@user`) and removed hashtag symbols (#).

[35] Mark Sedgwick, "Jihadist ideology, Western counter-ideology, and the ABC model," *Critical Studies on Terrorism* 5, no. 3 (2012).

[36] Tamim Al-Barghouti, The Umma and the Dawla: the nation state and the Arab Middle East (London: Pluto Press, 2008).

[37] A helpful analogy is to think of vectors as *points*. Two features X and Y make up a 2-D space. We can calculate the distance between 2-D points, i.e., $d = \sqrt{(X2-X1)^2 + (Y2-Y1)^2}$. Points closer to each other are more similar. More features make up an n-dimensional space, for which more sophisticated similarity functions exist besides the Euclidean distance function shown here.

## 4.1 In-domain evaluation

82% We trained[38] the LIBSVM machine learning algorithm[39] with a balanced training set of 45,000 HATE tweets (approx. 750,000 words) and 45,000 SAFE tweets. The predictive accuracy of the resulting model is **82%** (F1-score). To calculate the F1-score, we applied 3-fold cross-validation, meaning that we used a different 2/3 of the data for training and 1/3 for validation in three different tests and averaged the results.

Each test will yield true positives (actual HATE tweets predicted as HATE), true negatives (actual SAFE tweets predicted as SAFE), false positives and false negatives. which we can then use to calculate *recall* (TP / TP + FN) and *precision* (TP / TP + FP). Recall corresponds to *how many* hateful tweets we are able to expose, while precision corresponds to how many we can expose *without* falsely accusing anyone. For example, a system that flags every tweet as hate exposes all hate speech (high recall) but also calls everyone a jihadist (low precision).

| TEST 1 | TP | TN | FP | FN |
|---|---|---|---|---|
| HATE | 12725 | 11963 | 2887 | 2425 |
| SAFE | 11963 | 12725 | 2425 | 2887 |

| TEST 2 | TP | TN | FP | FN |
|---|---|---|---|---|
| HATE | 12533 | 12208 | 2832 | 2427 |
| SAFE | 12208 | 12533 | 2427 | 2832 |

| TEST 3 | TP | TN | FP | FN |
|---|---|---|---|---|
| HATE | 12490 | 12121 | 2989 | 2400 |
| SAFE | 12121 | 12490 | 2400 | 2989 |

Our model has a recall of 82.26% and a precision of 82.30%. The harmonic mean of both gives the F1-score. The error matrices of TP, TN, FP, FN for the three tests are shown on the right. The F1-score varies across languages, e.g., it is 79% for English, 84% for Arabic and 80% for French (Figure 11).

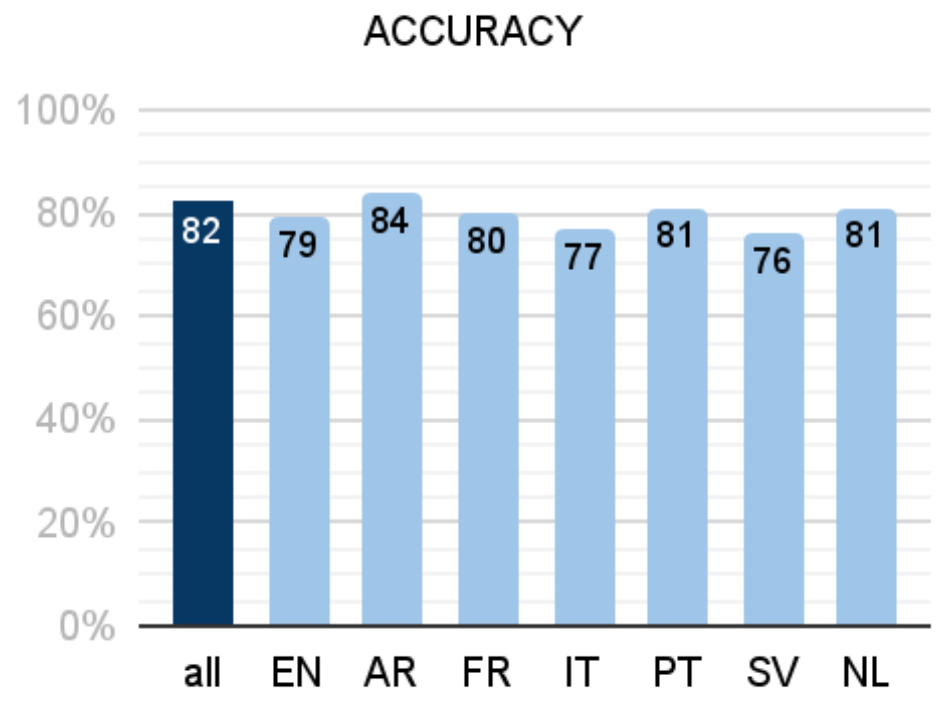

**Figure 11**. Accuracy by language.

While 82% accuracy is encouraging, the caveat is that our test was conducted under lab conditions: using an equal amount of HATE and SAFE tweets. Consequently, the chance of a correct random guess is 50%. In real-life, there will likely be thousands of SAFE tweets for every HATE tweet, making the problem much harder.

38 A demo in Python code with 250 HATE and 250 SAFE tweets can be found at: https://gist.github.com/tom-de-smedt/2d76d33f2515c5a52225af2bb4bb3900

39 Chih-Chung Chang and Lin Chih-Jen, "LIBSVM: A Library for Support Vector Machines," *ACM Transactions on Intelligent Systems and Technology* 2, no. 3 (2011): 27

Below are two examples of tweets not in the HATE corpus that are automatically detected as jihadist hate speech, using the trained system (one was still online in November 2017).

> RT @Abu___Suleiman: CHOKE ON YOUR FOOD: Murtadeen feeding Crusaders whilst thousands of Somali Muslim Muwahideen are facing famine!!!!!
> Posted by `@tawheedjihad_7` (ca. June 2015)

> RT @RidvanMemishi3: Muwahideen leaving in west and europe imediatly should leave, the world is going into chaos west is most propably dangerous area to live.
> Posted by `@umm_hawla__` (November 3, 2016)

## 4.2 Cross-domain evaluation

In Machine Learning, "domain adaptation" refers to the problem where a machine learning system appears to perform well on its own training and testing data (in-domain) but poorly on other related data (out-of-domain). We tested the performance of our trained system on the Kaggle ISIS dataset,[40] which has approximately 17,000 pro-ISIS tweets by 100+ authors. Only a handful of these authors (3) appear in our own corpus, making this a good scalability test. Our system correctly flags 76% of the Kaggle dataset as HATE. About 10% of the data consists of fairly neutral reporting (e.g., news updates by `@RamiAlLolah`), which the system correctly flags as SAFE, bringing the overall accuracy on the Kaggle dataset to 86%.

86% OUT-OF-DOMAIN LONG TEXTS

We then tested the performance on 50 jihadist manifestos from JustPaste.it, each containing about 2,500–4,000 words, written in Arabic, French and English (e.g., *Impediments of Takfir* by Al-Qaeda leader Asim Umar). These were mixed with a random sample of 450 Wikipedia articles (1:10 ratio) written in the same languages. The accuracy is 86% (40/50 of manifestos flagged as HATE, 390/450 of Wikipedia articles flagged as SAFE). In another out-of-domain test on long text, 93% of Lewis Carroll's novel *Alice in Wonderland* is flagged as SAFE.

80% OUT-OF-DOMAIN SHORT TEXTS

We also tested the performance on 5,000 messages collected from different jihadist Telegram channels, written in Arabic, English, French and Dutch (e.g., *Al-Fustaat Dutch*). In this case, the accuracy is 80%. The performance of our system is relatively stable across domains.

---

[40] Fifth Tribe, "How ISIS Uses Twitter," *Kaggle*, May 17, 2016, https://www.kaggle.com/fifthtribe/how-isis-uses-twitter

Finally, a related text analysis study[41] reports a preliminary 98% in-domain accuracy for a system trained on 5,000 English HATE and 3,000 SAFE tweets. This result is remarkable, but may have been prone to overfitting since the dataset is relatively small and spans only two months, i.e., the system might be memorizing the most frequent hashtags used during that time period instead of generalizing from linguistic patterns.

## 4.3 Industry applications

In the rapidly advancing field of Machine Learning, deep learning algorithms are nowadays favored, as they often yield higher accuracy, especially in combination with word embedding techniques. Unfortunately, Deep Neural Networks are also notoriously difficult to interpret. We have opted to use the more traditional Support Vector Machine (LIBSVM) to allow easy interpretation and reverse engineering of the results. Furthermore, our approach is able to process hundreds of texts in under a second and it can be trained in less than 10 minutes using off-the-shelf electronics, contrary to the processing power required by deep neural nets. This is advantageous in terms of deployment, since we can retrain with new data in real-time to stay ahead of the evolving rhetoric. We can think of 3 useful applications:

1. **Prevention.** The system is able to process large quantities of texts in real-time. It can be plugged into the pipeline of social networks such as Twitter to help stop the proliferation of online hate speech. Moderators would see a dashboard of recently flagged tweets up for inspection instead of having to manually sift through thousands of new tweets.
2. **Security.** The system is small (<100MB) as well as portable (e.g., PyInstaller) and could operate off a USB stick or on a Raspberry Pi. Law enforcement agencies could use it as a sorting algorithm to scan the hard drives of confiscated devices, to get an estimate of which text documents need inspection first.
3. **Analysis.** The system can be retrained quickly. It can be used as a discovery tool by scholars and Open-Source Intelligence (OSINT) personnel to monitor the evolution of jihadist hate speech rhetoric. While according to some media outlets ISIS has been defeated, the struggle between the resilient jihadist vision of a stateless unity of Muslims against the perverted and out-of-control Western culture is not over.[42]

---

[41] Michael Ashcroft, Ali Fisher, Lisa Kaati, Enghin Omer, and Nico Prucha, "Detecting Jihadist Messages on Twitter," in *2015 European Intelligence and Security Informatics Conference*, 161-164, 2015.

[42] Anthony N. Celso, "Zarqawi's Legacy: Al Qaeda's ISIS 'Renegade'," *Mediterranean Quarterly* 26, no. 2 (2015): 21-41.

## 4.4 Case study: Inside News

During February 2018, we monitored the *Inside News Global News Agency* profile on Twitter (`@InsideNewsAg`), which became active on January 29, 2018, steadily posting over 700 tweets to over 90,000 followers (Figure 12). On the surface, this profile posed as a neutral reporting agency, while our system flagged 70% of its tweets as suspicious.

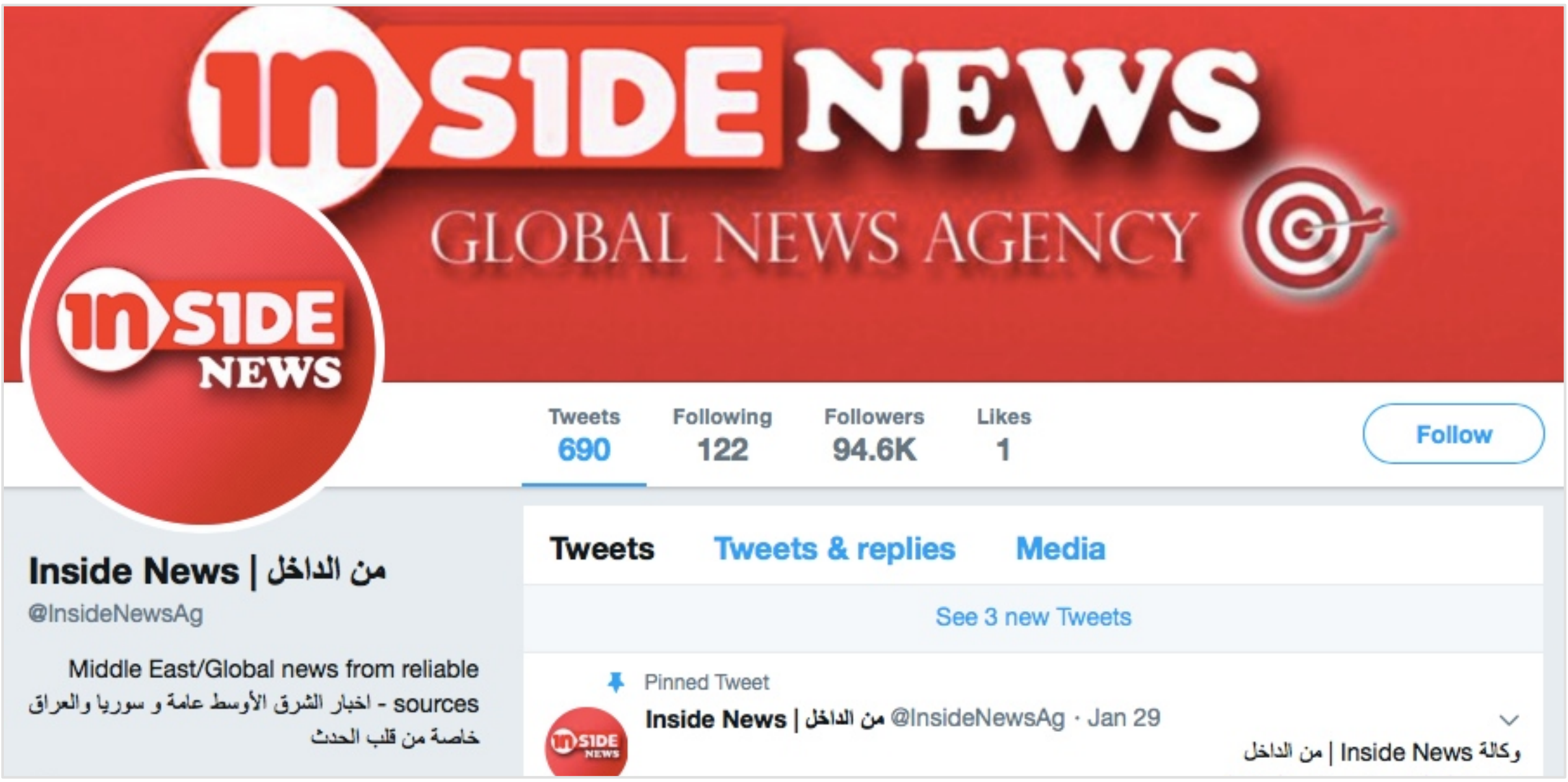

**Figure 12**. Inside News Agency employing countermeasures against identification.

Most tweets are not blatant hate speech, but contain more subtle hints of jihadist affiliation, such as sporadic references to "the apostates", once sharing a link to the *al-Naba* newsletter,[43] or distributing inside information with a biased, pro-ISIS undertone:

> Pro rebel channels spread the fake news that #IS soldiers threatened they will "rape" rebels women, to justify their continuous failue in #Yarmouk.
>
> Posted by `@InsideNewsAg` (February 17, 2018)

Most tweets are written in Arabic, others in near-perfect English. We discovered a number of idiosyncrasies in the author's English writing style, such as using *foto* instead of photo (7x), *camion* instead of truck (1x), and consistently quoting all proper names, leading us to believe that the author(s) might be of French or Moroccan origin. The profile was suspended from Twitter on February 28, 2018, but not before we identified 8,000 of the 90,000 followers.

---

[43] "New issue of The Islamic State's newsletter: 'al-Nabā' #120," *Jihadology*, http://jihadology.net/2018/02/22/new-issue-of-the-islamic-states-newsletter-al-naba-120

# 5 Network analysis

It has been noted that in-group radicalization within social networks (e.g., Twitter) has been key to radicalization to violence.[44,45] During data collection, we also collected the usernames of each of the followers (i.e., friends) of a Twitter profile and we monitored which profiles retweet (i.e., cite) what other profiles. This results in a set of 13,097 `X knows Y` relations and 21,373 `X cites Y` relations. Figure 13 shows a representation[46] of the `@Dabiq_Magazine` network, displaying users that cite `@Dabiq_Magazine` and users that in turn cite those users. The representation exposes a handful of clusters and an "information highway" in the center, but in general it is not very helpful because of its dense structure.

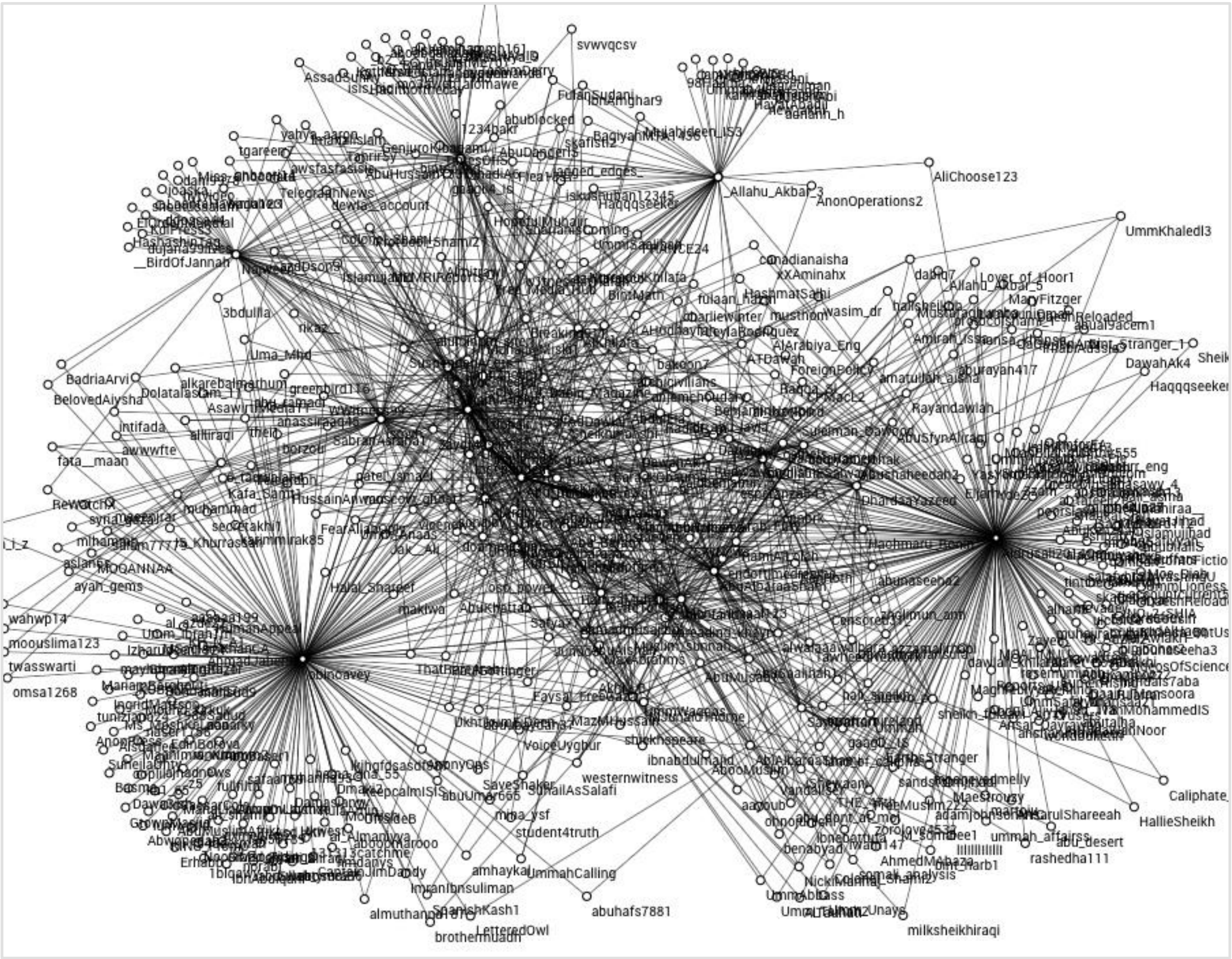

**Figure 13**. Twitter users that cite Dabiq Magazine and users that cite those users (ca. 2015).

---

[44] Pieter Van Ostaeyen, "Belgian Radical Networks and the Road to the Brussels Attacks," *CTC Sentinel* 9 (2016): 6.

[45] Marc Sageman, *Understanding Terror Networks* (University of Pennsylvania Press, 2004).

[46] Tom De Smedt. *Modeling Creativity: Case Studies in Python* (University Press Antwerp, 2013).

Graph Theory is a branch of mathematics that studies networks. It has produced efficient algorithms to identify the shortest paths between two points (i.e., nodes) in a network, to identify communities (i.e., clusters of nodes) and to identify important nodes. One technique called eigenvector centrality (cf. Google PageRank[47]) identifies the most important nodes by counting how many times they are referred to, taking into account how many times the referring nodes are in turn referred to. Such nodes are called *influencers* and correspond to users that control the flow of information.[48]

Using the Pattern toolkit, we applied eigenvector centrality > 25% as a filter to the Dabiq network to produce Figure 14. This representation is insightful: displaying the network of influencers that controlled the flow of information about Dabiq magazine around 2015.

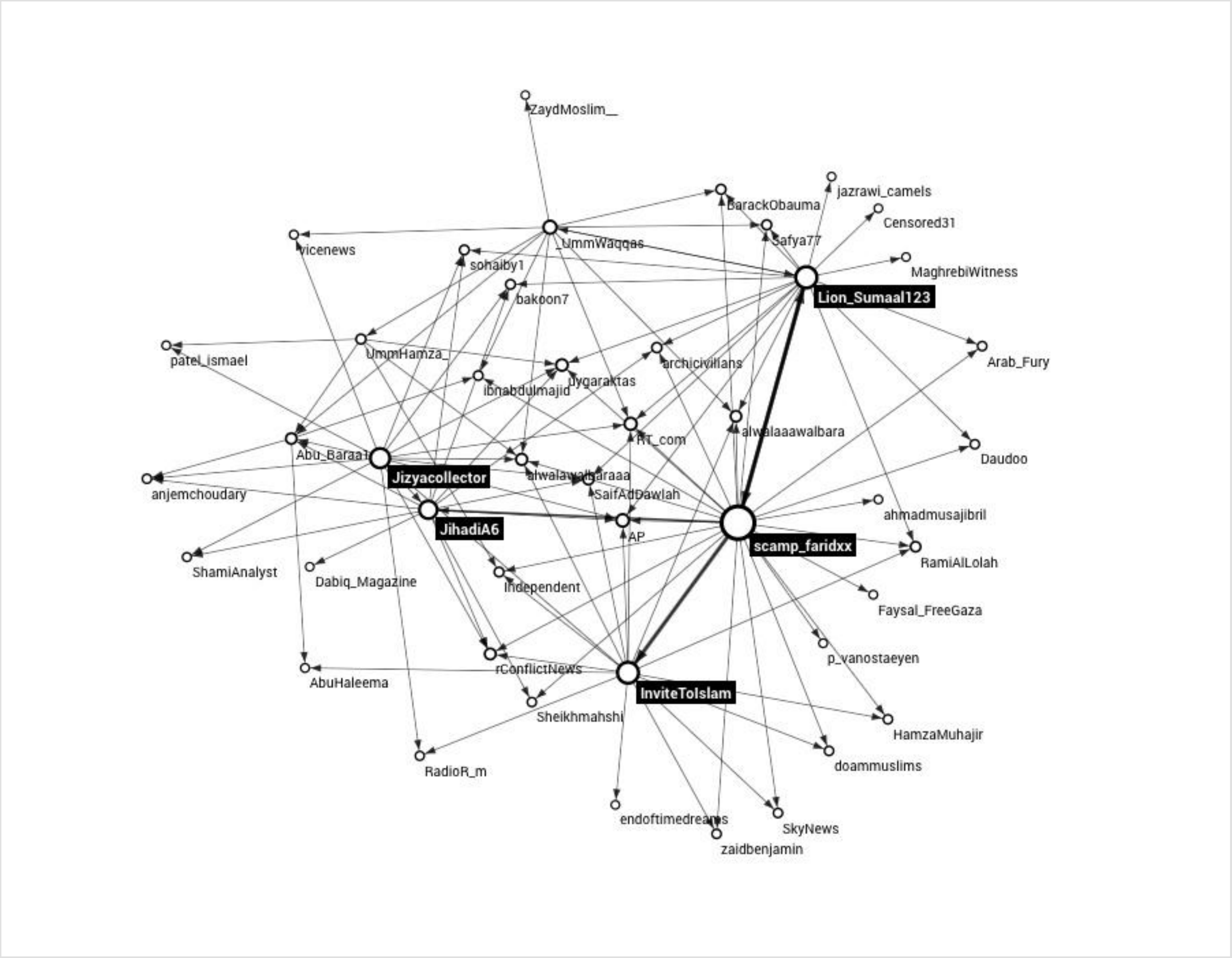

**Figure 14**. “Influencers” that cite Dabiq Magazine (ca. 2015).

---

[47] Lawrence Page, Sergey Brin, Rajeev Motwani, and Terry Winograd, “The PageRank Citation Ranking: Bringing Order to the Web” (Stanford University, 1999).

[48] Christine Kiss and Martin Bichler, “Identification of influencers – Measuring influence in customer networks,” *Decision Support Systems* 46, no. 1 (2008): 233-253.

In this representation, important nodes are bigger, and important information highways are broader. Node labels with eigenvector centrality > 50% have a black background. Several of the displayed profiles (e.g., `@anjemchoudary`) have indeed been prosecuted in 2016, suggesting that the technique has predictive merit. But manual review is always necessary, since the network may include false positives such as news agencies and experts (e.g., `@vicenews` and `@p_vanostaeyen`) that report their findings and in turn get cited by jihadists, as was the case with `@p_vanostaeyen`.[49] Finally, we wish to point out that a number of studies have also used more specialized methods to analyze terrorist networks, beyond our scope.[50]

# 6 Discussion

AI systems can be useful to assist with the detection of online jihadist hate speech. In our work, we have used automated statistical techniques with a stable predictive accuracy of 80%. The idea is not new: more than a decade ago, law experts already discussed the need for regulations and technology against online hate speech.[51] In more recent years, computer scientists have also discussed linguistic methods for the detection of hate speech, noting the absence of robust solutions.[52] More solutions are emerging, but online jihadist hate speech is a relatively new phenomenon, with as of yet no clear regulations or technological solutions. With this paper, we hope to contribute to the progressive insight. On request we will freely share our data with known intelligence agencies and research institutes.

# 7 Future work

We have attempted to automatically detect jihadist rhetoric by examining combinations of words, emoticons and spelling variations. However, present-day online communication involves not just text, but a combination of text, images, video, audio, and so on. Essentially, all communication is multimodal.[53] In future work, we want to examine how the automatic detection can be improved by combining text analysis with image recognition approaches on the set of 10,000 images we have collected.

---

[49] Bruno Struys, "Jihadsite versleuteld in strijd tegen terreur," *De Morgen*, January 12, 2017.

[50] Matthew C. Benigni, Kenneth Joseph, and Kathleen M. Carley, "Online extremism and the communities that sustain it: Detecting the ISIS supporting community on Twitter," *PLOS One* 12, no. 12 (2017).

[51] Alexander Tsesis, "Hate in Cyberspace: Regulating Hate Speech on the Internet," *San Diego Law Review* 38 (2001): 817.

[52] William Warner and Julia Hirschberg, "Detecting Hate Speech on the World Wide Web," on *Proceedings of the Second Workshop on Language in Social Media*, 19-26, 2012.

[53] Gunther Kress, Multimodality: A Social Semiotic Approach to Contemporary Communication (London: Routledge, 2009).